\documentclass[letterpaper, 10 pt, journal, twoside]{IEEEtran}

\IEEEoverridecommandlockouts                              %

\usepackage{graphicx} %
\usepackage{subcaption}
\usepackage{cuted}
\usepackage{booktabs}
\usepackage{multirow}
\usepackage{enumerate}
\usepackage{enumitem}
\usepackage{microtype}
\usepackage{algorithm}
\usepackage{algpseudocode}
\usepackage{cite}
\usepackage{siunitx}
\usepackage[breaklinks,colorlinks]{hyperref}
\usepackage[hang,flushmargin,symbol]{footmisc}

\usepackage{amsmath}
\usepackage{amssymb}
\usepackage{tensor}

\DeclareMathAlphabet{\mathcal}{OMS}{cmsy}{m}{n}

\def\N{\ensuremath{\mathbb{N}}}

\def\R{\ensuremath{\mathbb{R}}}

\def\F{\ensuremath{\mathcal{F}}}

\def\M{\ensuremath{\mathcal{M}}}
\def\N{\ensuremath{\mathcal{N}}}
\def\O{\ensuremath{\mathcal{O}}}

\renewcommand{\vec}[1]{\ensuremath{#1}}
\newcommand{\mat}[1]{\ensuremath{\mathbf{#1}}}
\newcommand{\norm}[1]{\left\lVert#1\right\rVert}

\newcommand{\prob}[1]{\ensuremath{p\left(#1\right)}}
\newcommand{\probc}[2]{\ensuremath{\prob{#1 \;\middle\vert\; #2}}}
\newcommand{\probdist}[2]{\ensuremath{p_{#1}\left(#2\right)}}
\newcommand{\probcdist}[3]{\ensuremath{\probdist{#1}{#2 \;\middle\vert\; #3}}}

\newcommand{\set}[1]{\ensuremath{\left\{#1\right\}}}
\newcommand{\fset}[2]{\ensuremath{\set{#1 \;\middle\vert\; #2}}}

\newcommand{\tf}[3]{\tensor[^{#1}]{\mat{#2}}{_{#3}}}

\DeclareMathOperator*{\argmin}{arg\,min}

\usepackage{xspace}

\newcommand{\ie}{\mbox{i.\,e.}\xspace}

\newcommand{\etal}{\emph{et al.}\xspace}
   
\renewcommand{\[}{\begin{equation}}
\renewcommand{\]}{\end{equation}}

\usepackage[capitalize]{cleveref}

\crefname{figure}{Fig.}{Figs.}
\Crefname{figure}{Figure}{Figures}
\crefname{section}{Sec.}{Secs.}
\Crefname{section}{Section}{Sections}
\Crefname{table}{Table}{Tables}
\crefname{table}{Tab.}{Tabs.}
\crefname{algorithm}{Algo.}{Algos.}
\Crefname{algorithm}{Algorithm}{Algorithms}
\crefname{appendix}{Sec.}{Secs.}
\Crefname{appendix}{Section}{Sections}

\title{Efficient Learning of Object Placement with Intra-Category Transfer}
\author{Adrian Röfer$^{1}$, Russell Buchanan$^{2,3}$, Max Argus$^1$, Sethu Vijayakumar$^2$, and Abhinav Valada$^1$
\thanks{This paper was recommended for publication by
Editor Aleksandra Faust upon evaluation of the Associate Editor and Reviewers’
comments.}
\thanks{\noindent This work was funded by the BrainLinks-BrainTools center of the University of Freiburg, the H2020 Project HARMONY, and the Alan Turing Institute. Adrian Röfer acknowledges travel support from the EU H2020 research and innovation programme under grant agreement No 951847.}
\thanks{Manuscript received: November, 5th, 2024; Revised February, 7th, 2025 \& August, 12th, 2025; Accepted November, 4th, 2025.}
\thanks{\textit{Corresponding Author}: Adrian Röfer, \textit{email}: aroefer@cs.uni-freiburg.de.}
\thanks{$^{1}$ Department of Computer Science, University of Freiburg, Germany. %
}
\thanks{$^2$ School of Informatics, University of Edinburgh, Scotland.
}
\thanks{$^3$ Faculty of Engineering, University of Waterloo, Canada.
}
\thanks{Digital Object Identifier (DOI): 10.1109/LRA.2025.3632727.}
\thanks{© 2026 IEEE.  Personal use of this material is permitted.  Permission from IEEE must be obtained for all other uses, in any current or future media, including reprinting/republishing this material for advertising or promotional purposes, creating new collective works, for resale or redistribution to servers or lists, or reuse of any copyrighted component of this work in other works.}
}

\newcommand{\scenescorest}{$\text{SS}^{16}$}

\newcommand{\changecol}{\color{black}}
\newcommand{\textcol}{\color{black}}
\newcommand{\change}[1]{\changecol #1\textcol}

\makeatletter
\g@addto@macro{\endtabular}{\rowfont{}}%
\makeatother
\newcommand{\rowfonttype}{}%
\newcommand{\rowfont}[1]{%
\gdef\rowfonttype{#1}#1\ignorespaces
}

\newif\ifmix
\mixtrue %
\newif\arxiv

\begin{document}
\bstctlcite{IEEEexample:BSTcontrol} %
\maketitle

\begin{abstract}
    Efficient learning from demonstration for long-horizon tasks remains an open challenge in robotics. While significant effort has been directed toward learning trajectories, a recent resurgence of object-centric approaches has demonstrated improved sample efficiency, enabling transferable robotic skills. Such approaches model tasks as a sequence of object poses over time. In this work, we propose a scheme for transferring observed object arrangements to novel object instances by learning these arrangements on canonical class frames. We then employ this scheme to enable a simple yet effective approach for training models from as few as five demonstrations to predict arrangements of a wide range of objects including tableware, cutlery, furniture, and desk spaces. We propose a method for optimizing the learned models to enable efficient learning of tasks such as setting a table or tidying up an office with intra-category transfer, even in the presence of distractors. We present extensive experimental results in simulation and on a real robotic system for table setting which, based on human evaluations, scored 73.3\% compared to a human baseline. We make the code and trained models publicly available at \url{https://oplict.cs.uni-freiburg.de}.
\end{abstract}


\IEEEpeerreviewmaketitle

\section{Introduction}
\label{section:intro}

\IEEEPARstart{H}{umans} excel at teaching each other various skills efficiently. Whether it is setting tables, changing bicycle brakes, or furnishing a room, we can guide another person to proficiency within a handful of training sessions. Deep learning on large datasets has yielded impressive manipulation results on a variety of tasks~\cite{Bahl2023,chi2024diffusion,chisari2024learning}. However, the requirement for vast amounts of data by these methods prevents them from achieving human-like efficiency in skill acquisition~\cite{chisari2022correct}. %

One of the reasons for this requirement is the need to distill relevant features from visual observations. Pre-trained vision foundation models that successfully transfer across different object instances have alleviated this problem. They enable one- or few-shot trajectory imitation by using deep vision features for object-centric localization and low-capacity models for encoding motion~\cite{gao2023kvil,heppert2024ditto,von2023treachery}. Another method for overcoming the data requirement is the sparsification of tasks. Subsampling the number of timesteps in motion policies can boost the efficiency of reinforcement-learning~\cite{nematollahi2022sacgmm}. Taken to its extreme, such a sparsification yields a separation of tasks into discrete sequences of motions, with boundary conditions for transitioning between segments~\cite{toussaint2022sequence,wang2023temporal,merlo2025exploiting}. In this view, instead of learning a full trajectory for placing a monitor on a desk, the agent learns where a monitor should go in relation to keyboard and mouse. The specific motion is then to be inferred at runtime. 
Such sparse representations have been used successfully in teaching robots simple pick-and-place sequences from a single demonstration~\cite{guo2023learning,zhu2024vision}, while it has also been possible to learn control sequences~\cite{mao2023learning} for the motion segments.

\begin{figure}
    \centering
    \includegraphics[width=0.8\linewidth]{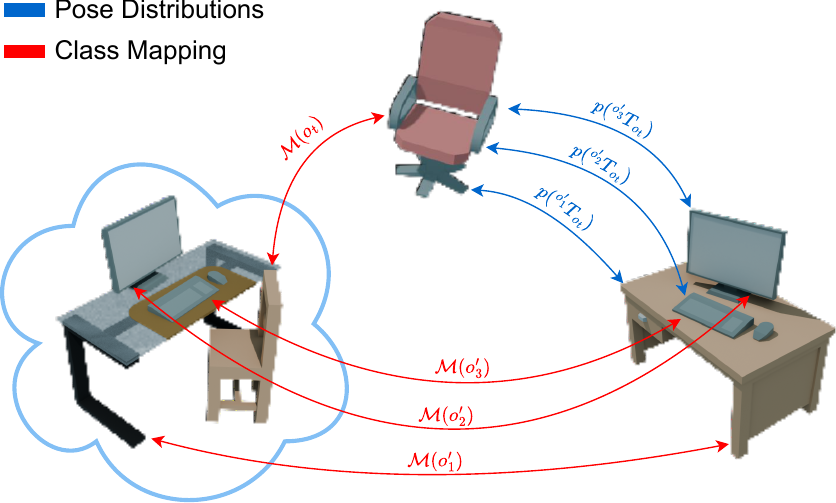}
    \caption{%
        Our approach learns object placements sample-efficiently by mapping object instances to a known canonical instance and inferring the placement of the new object in this canonical space. Here, the known setup on the left is matched with the novel one on the right to place the chair. %
    }
    \label{fig:teaser}
    \vspace{-0.5cm}
\end{figure}

\begin{figure*}
    \vspace{4mm}
    \centering
    \includegraphics[width=0.9\textwidth]{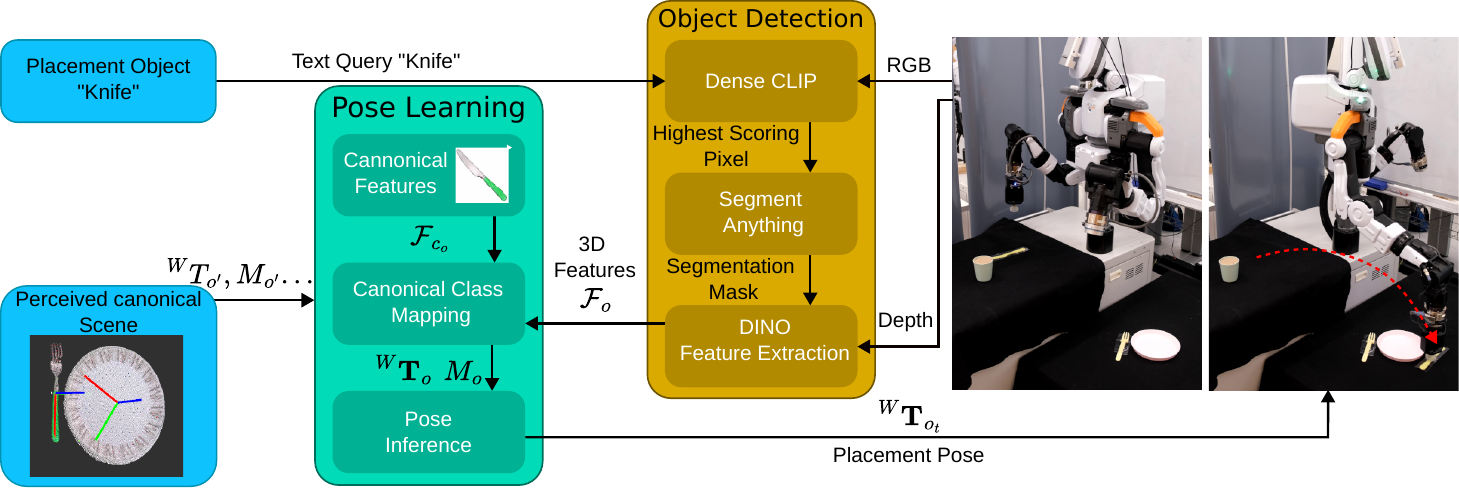}
    \caption{Full pipeline of our system with real robot experiment. Our proposed pose inference method predicts the ideal object placement pose, which the robot then arranges autonomously. Our approach's few-shot transfer to other object instances is enabled by our object class mappings which are enabled by several large networks for object detection and feature extraction.}
    \label{fig:pipeline}
    \vspace{-0.5cm}
\end{figure*}

In this work, we aim to support long-horizon task learning by proposing a method for learning \textit{where} objects should be during the \textit{key moments} of sequential placement task. We introduce an approach for learning relative arrangements of objects which is trainable using a minimal number of examples ($\leq 5$) and transfers novel object instances of the same classes as seen during training.
We employ this approach to incremental object placement tasks in which larger everyday scenes of objects have to be arranged. Such natural scenes, similar to \cref{fig:teaser}, can exhibit complex inter-object dependencies and significant object variability which creates a challenge to reproducing the scenes in a semantically acceptable manner. %
Our framework operates on distributions of relative poses, for which we introduce a scheme for mapping object observations to a canonical class frame, which enables the desired intra-category transfer. In addition, we introduce a method for minimizing the complexity of models and reducing the impact of spurious correlations formed with unrelated objects. We demonstrate our method in simulations, showing the effectiveness of our method by arranging furniture, tableware, and office space. In simulated experiments, our method was successfully trained with as few as five demonstrations and performed robustly against distractors. Using the system depicted in~\cref{fig:pipeline}, we perform real robot experiments on tableware arrangements. Our system was rated by human evaluators as $73.3\%$ as good as human performance. Our primary contributions are:
\begin{enumerate}[topsep=0pt,noitemsep]
    \item A novel approach for few-shot relative pose learning.
    \item A framework for mapping observations of objects to a canonical class frame for intra-category transfer.
    \item A technique for optimizing model complexity and removing distracting observations from the models.
    \item Real-world robot experiments of autonomously setting tables using both familiar and unfamiliar objects, with the quality of the arrangements assessed by human jurors.
    \item We publicly release our dataset, code, and models at \url{https://oplict.cs.uni-freiburg.de}.
\end{enumerate}

\section{Related Work}
We briefly summarize related work on learning manipulation tasks and learning relative object poses.

\noindent\textit{Manipulation as a Series of Contacts}:
Manipulation tasks can be understood as a series of contact state changes instead of dense temporal trajectories. Recent works~\cite{guo2023learning,zhu2024vision} have demonstrated that it is possible to learn a long-horizon manipulation task from a single demonstration by using changes in contacts as delimiters of robotic actions. Similarly, Mao~\etal~\cite{mao2023learning} demonstrate that such a delimitation can be used to simplify the combinatorial problem of searching for effective action sequences in long-horizon tasks, with learned parameterizations of actions and success predictors for action primitives and their sequences. In motion planning, this sparse segmentation identifies \emph{kinematic modes} under each of which manipulation is continuous, making it efficiently solvable using numerical optimization~\cite{toussaint2022sequence}. %
Despite the sparsity, contact-centric representations are also informative as demonstrated by \cite{aksoy2011learning} who introduced a representation of tasks as a sequence of contact graphs between objects, which they coined \emph{semantic event chains}, and could recognize objects and tasks from this representation. In \cite{aksoy2011execution}, they illustrated how their representation can support manipulation, but the focus is on recognition of tasks and objects~\cite{toumpa2023object,ziaeetabar2023hierarchical,sochacki2023recognition}.%

\noindent\textit{Object Pose Learning}: 
While several works investigate object pose placement from language cues~\cite{kim2024lingo,zhai2023sg,gkanatsios2023energy}, only a few learn relative object placement directly from category information, without any cues.
Image-space approaches are conditioned on language instructions to predict image-space activations for object placement~\cite{mees2020learning}. More recently~\cite{simeonov2023shelving} demonstrated an approach for finding arrangement poses of single objects in cluttered scenes using diffusion processes on point cloud data. In~\cite{liu2022structdiffusion,yang2023diffusion}, the authors demonstrate approaches that learn language-conditioned rearrangement of objects. These approaches incrementally update a given scene towards the described arrangement until the updates anneal. In~\cite{kartmann2023interactive,kim2024lingo}, object-placement or navigation locations are identified from verbal descriptions. Moreover, recent methods also use pretrained VLMs, or generative models, to generate desired arrangements using language cues with zero-shot transfer. Methods such as~\cite{kapelyukh2024dream2real,wu2023tidybot} either generate desired target arrangements or query pretrained language models for target locations for objects. Although the knowledge that can be extracted from these models is impressive, they ultimately require both very specific prompts and very good observability of a scene. In contrast, our aim in this work is to solely leverage data obtainable from physical demonstrations without language cues, even without full observability.

The most similar works to our own are~\cite{kapelyukh2022my,kapelyukh2023scenescore,welschehold2019combined}, which also do not use language cues. 
Kapelyukh~\etal~\cite{kapelyukh2022my} employ graph-neural networks to predict object arrangements from user preferences extracted from demonstrations. They represent the objects in a scene using text embeddings and their positions and train an embedding VAE on the fully connected graph of a scene. They can infer a user's preference from a scene the user arranges and use their model to rearrange other scenes to the user's liking. Their follow-up work~\cite{kapelyukh2023scenescore} proposes a model that assigns a score to a given scene, again using a graphical representation and CLIP embeddings~\cite{radford2021learning} of objects. Unlike our approach, these methods require 16 or more training examples, while 5 are sufficient for ours. Their inferences are limited to SE2, while we address full SE3 poses. %
Most similar to our approach is the \textit{Intention Likelihood} formulation~\cite{welschehold2019combined}. It also employs relative pose observations and can be trained with as few as 5 demonstrations. Distinctively, it does not learn a distribution over poses but assumes a fixed kernel density and scores poses under it. At inference time, the method places previously seen poses in the scene and scores their likelihood. Our method estimates the density of the pose distributions from the training data and samples these, making our inferences more precise and less sensitive to distractors.

\section{Problem Definition}
As a multi-step task with complex inter-object dependencies, we study the problem of sequential object placement, which assumes that a sequence of objects $\O_p = \set{o_1, \ldots, o_n}$ needs to be placed in a scene in relation to a static set of objects $\O_s$. Each placed object transitions from $\O_p$ to $\O_s$. Formally: At time step $t = 1$, the sets are the initial sets $\O_p$ and $\O_s$, afterwards they change to $\O_{p,t} = \set{o_t, \ldots, o_n}$ and $\O_{s,t} = \O_s \cup \set{o'_1, \ldots, o'_{t-1}}$ for all $t \leq n$.

Each object $o$ has a world-space pose $\tf{W}{T}{o}$ and a class $c_o$. Each class is associated with a set of feature points, $\F_c = \set{(e_1, p_1), \ldots, (e_m, p_m)}$ which consist of an embedding $e \in \R^E$ and a position $p \in \R^3$ in a canonical class frame. The generation of feature points and their utility for manipulation and pose estimation and tracking is widely studied in robotics.

Given this structure, we seek to learn the distribution of object placements in the scene, given its class, and the poses and classes of the other already placed objects: $\probcdist{t}{\tf{W}{T}{o_t}}{c_{o_t}, c_{o'_1}, \ldots, c_{o'_{t-1}}, \tf{W}{T}{o'_1}, \dots \tf{W}{T}{o'_{t-1}}}$, with $o'_1, \ldots, o'_{t-1} \in \O_{s,t}$.

\section{Pose Learning Approach}

In order to learn poses effectively, we decompose of learning world space poses $\probcdist{t}{\tf{W}{T}{o_t}}{\ldots}$ into 
\[
\label{eq:joint_prob}
\begin{aligned}
\probcdist{t}{\tf{W}{T}{o_t}}{c_{o_t}, c_{o'_1}, \ldots, c_{o'_{t-1}}, \tf{W}{T}{o'_1}, \dots \tf{W}{T}{o'_{t-1}}} &= \\\prod_{o' \in \O_{s,t}} \probcdist{t}{\tf{o'}{T}{o_t}}{c_{o_t}, c_{o'}},
\end{aligned}
\]
learning of relative poses $\tf{o'}{T}{o}$, which we derive as $\tf{o'}{T}{o} = \tf{W}{T}{o'}^{-1}\cdot\tf{W}{T}{o}$. To enable intra-category transfer, we assume there exists an invertible class mapping $\M(o) = o_c$ which maps an object's properties, \ie its pose and feature points, to the canonical categorical instance. These concepts, depicted in \cref{fig:teaser}, yield the probability of a relative pose given the observed distribution of poses in categorical space as
\[
\label{eq:single_prob}
\begin{aligned}
    \probcdist{t}{\hspace{-0.5mm}\tf{o'}{T}{o_t}\hspace{-0.5mm}}{\hspace{-0.5mm}c_{o_t}, c_{o'}\hspace{-0.5mm}} &= \probc{\hspace{-0.5mm}\tf{o'}{T}{o_t}\hspace{-0.5mm}}{\hspace{-0.5mm}\M(o')\tf{o'}{T}{o} \M(o_t)\hspace{-0.5mm}}. %
\end{aligned}
\]

\subsection{Canonical Class Mappings}

Class maps deform an instance of an object to best match a known canonical object. Our approach requires these maps to be affine transformations, but in this work, we only consider linear scaling instances. The simplest considered map is the identity $\M_I = I$, which does not scale an object to its categorical representation.
The \emph{uniform linear map}, which was used in \cite{goodwin2023you} applies a single scaling factor $s$, to scale the observed instance of the object to the class prototype. We denote the map as 
\[
\M_U(s) = diag(s, s, s, 1).
\]
Given an observed instance $o$ and its feature points $\F_o = \set{p_1, \ldots, p_m}$, we can derive the scaling factor $s$ easily, by comparing the distances between point pairs in $\F_o$ and $\F_{c_o} = \set{\hat{p}_1, \ldots, \hat{p}_m}$ as
\[
s = \frac{1}{|F_o|^2} \sum_{i=1}^m \sum_{j=1}^m \frac{\norm{\hat{p}_i - \hat{p}_j}}{\norm{p_i - p_j}}.
\]
We find this mapping to perform robustly, as it is simple to derive. Nonetheless, its assumption of a single scaling factor is limiting, as objects can vary quite significantly and non-uniformly in their extents, \ie slender wine glasses compared to bulbous ones. For such cases, we propose the generalization $\M_O$ of the previous mapping as:
\[
    \M_O(s_x, s_y, s_z) = diag(s_x, s_y, s_z, 1),
\]
which we refer to as \emph{orthogonal linear map}. Deriving the values for $s_x, s_y, s_z$ is more challenging in this case, as the extents have to be measured in the object's frame $\tf{W}{T}{o}$, while the estimate of the scaling factors also affects the estimation of this frame. Thus, we jointly optimize fit and pose as
\[
\min_{\tf{W}{T}{o}, \vec{s}} \sum_{i=0}^m \norm{\vec{p}_i - \M_O(s_x, s_y, s_z)^{-1} \tf{W}{T}{o} \hat{\vec{p}}_i},
\]
where $\M_O(s_x, s_y, s_z)^{-1}$ is the inverse of the class mapping, thus mapping the class' feature points to the space of the observed object instance. 
With this estimate of $\M_O$, we attempt a better fit of $\tf{W}{T}{o}$. This process continues until it converges or until a fixed step limit is reached.

\subsection{Learning Relative Pose Distributions}
\label{sec:approach_rel_poses}

To learn the relative poses of objects in category space, we use multivariate normal distributions. Our distributions capture the relative pose $\tf{b}{\hat{T}}{a}$ of two objects $a, b$ where
$\tf{b}{\hat{T}}{a} = \M_b \tf{b}{T}{a}$, with $\M_b$ being the class mapping for the object $b$.
While $\tf{b}{\hat{T}}{a} \in \R^{4\times4}$ is a convenient representation for computations, it does not lend itself to learning due to its size and redundancy. Instead, we represent these poses in a lower-dimensional feature space. We represent the encoding into this space as an invertible function $f$. These distributions capture the conditional probability defined in \cref{eq:single_prob}:
\[
    \probcdist{t}{\tf{b}{T}{a}}{c_{a}, c_{b}} = \prob{f(\tf{b}{\hat{T}}{a})}.
\]
We use these pair-wise distributions to learn one joint placement distribution for each $o \in \O_{p}$. The naive approach is the substitution of the pair-wise conditional probability in \cref{eq:joint_prob} for our newly derived pair-wise probability. We refer to this unidirectional model as $p_{U, t}$.
A problem with this model is its disregard for the placed object's geometry, as only the reference object's class mapping is used. We propose a bidirectional model $p_{B, t}$ which includes the observation of the reference object:
\[
\probcdist{B, t}{\tf{o'}{T}{o_t}}{c_{o_t}, c_{o'}} = \prob{f(\tf{o'}{\hat{T}}{o})} \prob{f(\tf{o}{\hat{T}}{o'})}. %
\]
The full model for placing one object given all other objects in the scene can then be derived by substituting \cref{eq:joint_prob}. For inference on either model, we sample uniformly from $o' \in \O_{s,t}$ and draw a sample $\vec{x}$  from its relative distribution. We use $\tf{W}{T}{x} = \tf{W}{T}{o'} (M_{o'}^{-1} f^{-1}(x))$ to compute its corresponding pose and then score the sample under the probability density function of our joint distribution $p_t$. Naively, we pick the highest-scoring sample. Experimentally we observed that the average sample $\hat{x}$ of the top 10 samples $X_{10}$ almost always scored better than each individual sample, \ie $\forall x \in X_{10}: p_t(\hat{x}) \geq p_t(x)$. We exploit this in a simple refinement scheme: We draw and initial $10^5$ samples, take the top 10 samples according to $p_t$, and form a new distribution of world-space poses with mean $\hat{x}$ and variance $var(X_{10})$, represented as position and rotation vector. We sample a new $10^3$ from this distribution and repeat the process until we see only minor increases in $p_t$ or until a step limit is reached.

\subsection{Pose Encoding \& Model Minimization}
\label{sec:encoding_choice}

The choice of pose-encoding $f$ affects the type of relative relationship which can be captured by $\prob{\tf{o'}{T}{o}}$. In this work, we examine a number of choices for $f$ and compare their general utility for inference in \cref{sec:sim_experiments}. %
Instead of choosing a fixed encoding for all relations, we would like to be able to autonomously identify the ideal pose representation for a given set of relative pose observations. To do so we propose selecting the encoding $f_{t,o',o}$ for these observations as 
\[
    f_{t,o',o} = \argmin_{f \in F} H(\N(\mu(f(\tf{o'}{T}{o})); \Sigma (f(\tf{o'}{T}{o})))),
\]
where $H$ denotes the Shannon information of the distribution fitted to the relative pose observations encoded using $f$. We use the specialized entropy for multivariate Gaussians $H=\frac{1}{2} \ln ((2\pi e)^k \det \mat{\Sigma})$. Assuming that two encodings $f, g$ do not produce vastly differently scaled vector spaces, we can use this criterion to evaluate the tightness of the model's fit.

With the described method, we can optimize the choice of observation encoding. As we described the method so far, we form distributions for all relative object observations in a scene. Especially in larger scenes, many of the relative pose observations will not be relevant to the placement of an object. Imagine placing a cup on a coffee table that is set in front of a sofa. While the relative pose of cup and sofa will have a statistical trend, semantically, this correlation is (largely) spurious. When performing inference, maintaining all of these relationships potentially increases the number of samples needing to be drawn and the numerical instability.

\begin{figure}
    \vspace{1mm}
    \centering
    \includegraphics[width=\linewidth]{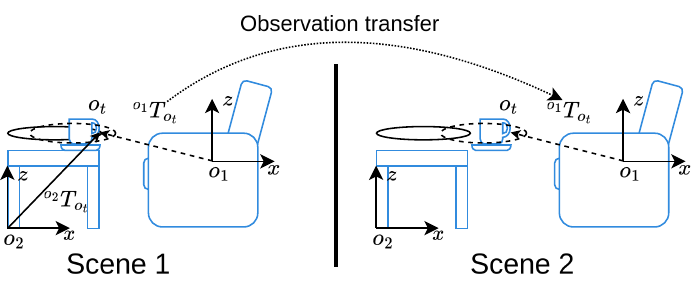}
    \caption{Illustration of the observation augmentation procedure used for model pruning. The observation of the cup relative to the sofa in Scene 1 is transferred to Scene 2. While the transferred observation is scored the same from the point of the sofa, from the point of the table, it is scored far lower, informing us that including the table improves our model.
    }
    \label{fig:outlier_augmentation}
    \vspace*{-0.3cm}
\end{figure}

\begin{figure*}
    \centering
    \includegraphics[width=0.19\textwidth]{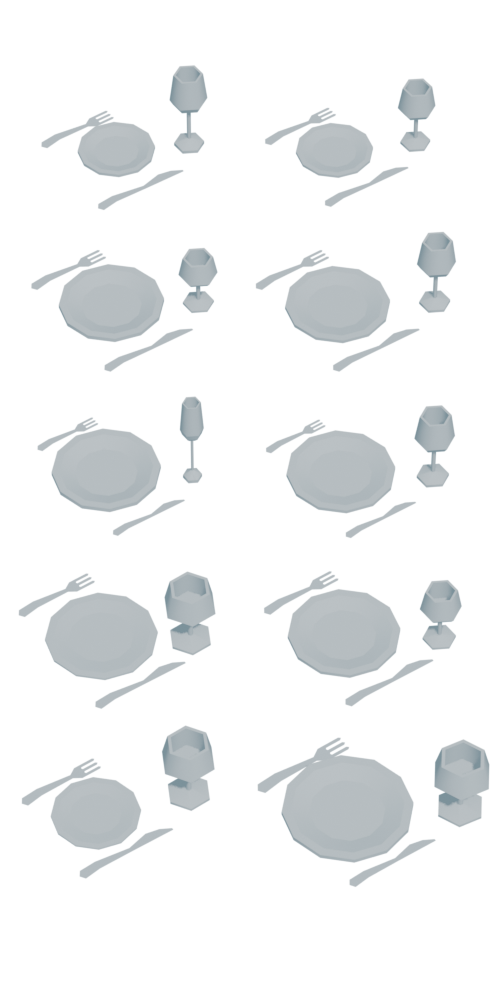}
    \includegraphics[width=0.19\textwidth]{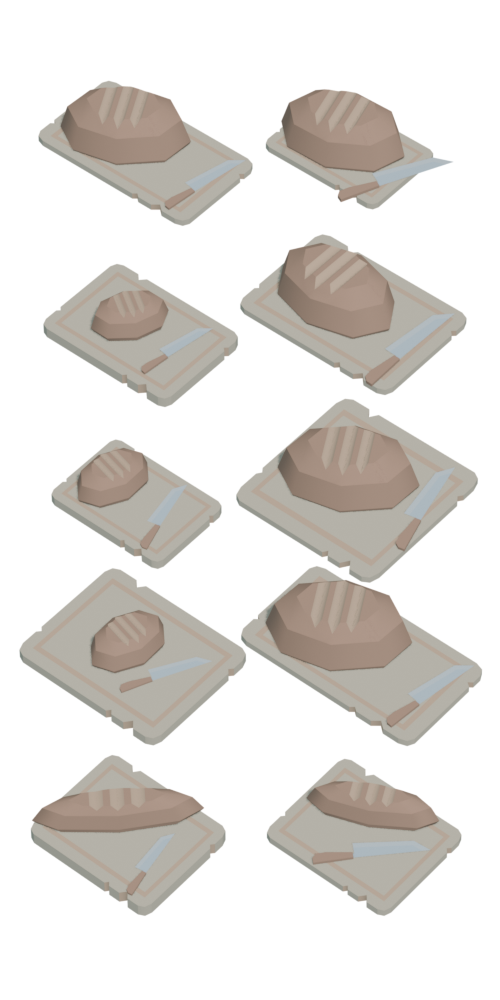}
    \includegraphics[width=0.19\textwidth]{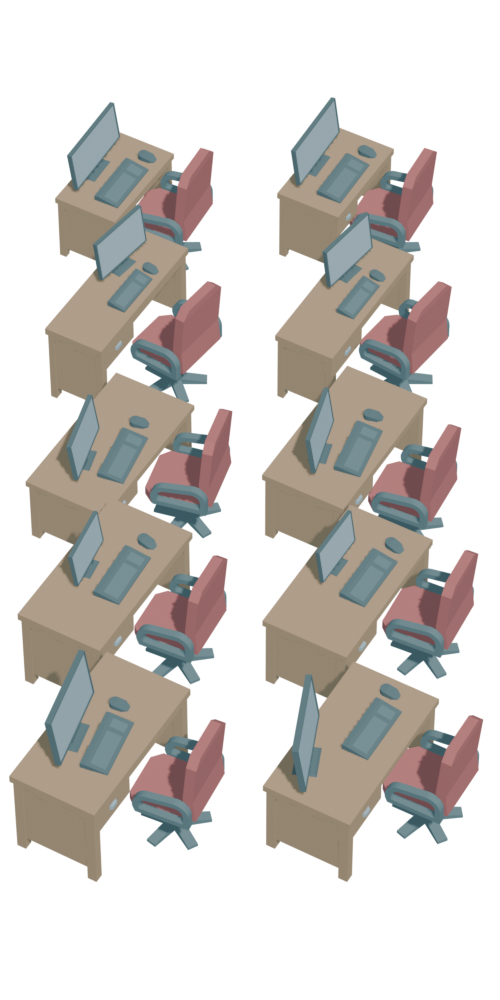}
    \includegraphics[width=0.19\textwidth]{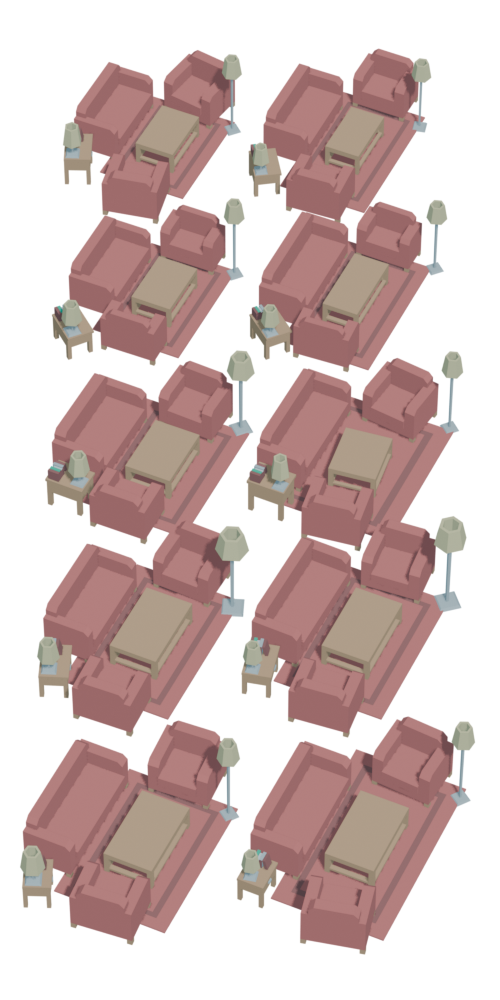}
    \includegraphics[width=0.19\textwidth]{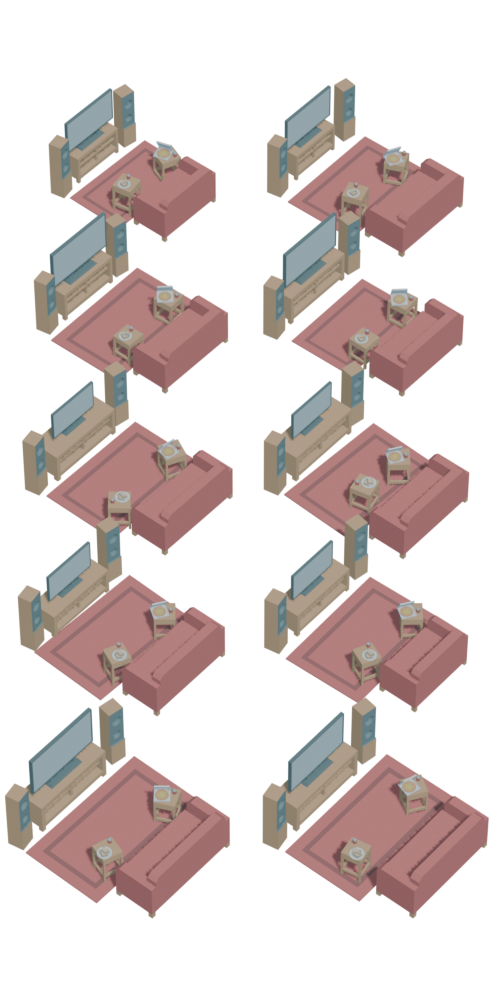}
    \caption{We evaluate five different training scenarios with 16 variations for training (only 5 shown here) and 5 for evaluation each. The scale of objects changes non-uniformly and the placements are varied. They are hand-crafted to ensure that they are semantically meaningful. For each object category, we generate 12 key points, which remain the same across all instances. From the top left: \textit{Dinner places}, \textit{Bread-cutting}, \textit{Desks}, \textit{Living room}, and \textit{TV setup}.}
    \label{fig:fulldataset}
    \vspace{-0.5cm}
\end{figure*}

Thus, given the fitted distributions $\probdist{o_1}{\tf{o_1}{T}{o_t}}, \ldots, \probdist{o_{t-1}}{\tf{o_{t-1}}{T}{o_t}}$, we seek to identify the minimal set of distributions $O^*_{s,t} \subseteq \O_{s,t}$ which represents the data. We propose doing so by using an outlier-discrimination strategy: We can expect all training samples $\tf{W}{T}{o_{t}}^k$ to be reasonably probable under the fitted distributions in the originally observed context of scene $k$. However, we can produce potential outliers by taking observation $\tf{o_1}{T}{o_{t}}^i$ from scene $i$ and introducing it into scene $j$ as $\tf{W}{\tilde{T}}{o_t}^j = \tf{W}{T}{o_1}^j \cdot \tf{o_1}{T}{o_t}^i$. We can now evaluate $y_j = \probdist{o_2}{\tf{o_2}{T}{W}^j \cdot \tf{W}{\tilde{T}}{o_t}^j}$ and compare it to $y_i = \probdist{o_2}{\tf{o_2}{T}{o_t}^i}$. If $p_{o_1}$ and $p_{o_2}$ identify largely overlapping regions, \ie they represent redundant information, we would expect $y_i \approx y_j$. If the distributions are not redundant, we would expect $y_i \gg y_j$. \cref{fig:outlier_augmentation} illustrates this process. We use a sampling-based scheme to incrementally build $O^*_{s,t}$. 
We initialize this set as $O^*_{s,t} = \set{o'_1}$ where $o'_1$ is sampled from $\O_{s,t}$ according to $\prob{o' \in O^*_{s,t}} \propto H(\tf{o'}{T}{o_t})$. Given this root object, we generate $K^2$ potential outlier observations. For each object $o'_i \in \O_{s,t}$ with $o'_i \not = o'_1$, we compute $y_{i,k,j}$. We define the set of rejected samples of an object $o$ as
\[
    R(o) = \fset{(k,j)}{\frac{y_{o,k,j}}{y_{o,k,k}} < \alpha},
\]
and the rejected samples of a set of objects as $R(O) = \bigcup_o^O R(o)$.
We now incrementally expand $O^*_{s,t}$ by sampling an object $o' \in \O_{s,t} / O^*_{s,t}$ with $\prob{o'} \propto |R(o) / R(O^*_{s,t})|$. For a selected sample, we calculate a score $s_{o'}$ as
\[
    s_{o'} = \frac{|R(o') / R(O^*_{s,t})| \cdot \tilde{H}(O^*_{s,t})}{(\hat{K}- |R(O^*_{s,t})|) ( \tilde{H}(O^*_{s,t}) - \tilde{H}(O^*_{s,t} \cup \set{o'}))},
\]
where $\hat{K} = K^2-K$ and $\tilde{H}(O) = \frac{1}{|O|} \sum_{o\in O} H(p(\tf{o}{T}{o_p}))$ is the mean entropy of the distributions of a set. Intuitively, this score trades off the fraction of remaining outliers with the relative rise in mean entropy of the chosen set. We compare $s_{o'}$ to a random sample $\epsilon \in [0, 1]$ and admit $o'$ to $O^*_{s,t}$ if $s_{o'} > \epsilon$. Otherwise, we terminate the assembly of $O^*_{s,t}$. We repeat this process multiple times and pick the best-sampled model according to scoring lowest under $(1 - |R(O)|/\hat{K}) \cdot \tilde{H}(O)$.
To limit the computational cost of generating hypothetical scenes, we prefilter a subset $\hat{\O}_{s,t} \subseteq \O_{s,t}$, where $\prob{o \in \hat{\O}_{s,t}} \propto H(p_o)$ and $|\O_{s,t}| \propto K$ from which we select objects. The complexity of this process is proportional to $|\O_{s,t}|K^2$.

\section{Experimental Evaluation}

We evaluate our approach both in simulation and on a real robot. First, in simulation, we measure the impact of feature encodings, category maps, relative pose models, impact of training samples, and impact of distractor objects. Second, we deploy the best performing model on a real robot and use it for table setting with seen and unseen cutlery.

\subsection{Simulation Experiments}
\label{sec:sim_experiments}
For the simulated experiments, we use the five scenes shown in \cref{fig:fulldataset}. We train our models with different feature and category maps for predicting the placement poses of the objects with predetermined placement order. We evaluate model performance in $n$-step inference, where $n$ truncates the length of the task from the back, starting at the final placement step $t$. This is best understood in reference to the placement sequence of object $\O_p$: The $n$-step inference of object $o_t$ reduces $\O_p$ to $\hat{\O}_p = \set{o_{t-n}, \ldots, o_t}$, with the initial observation set $\O_{s,t-n}$. For example, $0$-step inference is the placement of the current object, where previously placed objects have ground truth pose. As $n$ increases, the error in the scenes accumulates for each placed object. 

We evaluate the performance of different combinations of category maps $\M$, pose representations $f$, models $p_U, p_B$, and the impact of model minimization by fitting our models to $5$ training samples and evaluating on $5$ test scenes. Our choices for pose encodings are $f_{quat}, f_{AA}, f_{euler}$ which all encode poses as Cartesian position and rotations as a quaternion, rotation vector, and Euler angles, respectively. The $f_{\mathfrak{se3}}$ encoding uses the log-map to represent a pose as an element of $\mathfrak{se3} \subset \R^6$. The $f_{mix}$ encoding chooses between the aforementioned ones according to the heuristic described in~\cref{sec:encoding_choice}.
We start by comparing the impact of combinations of $\M$ and $f$ and the performance of our two competing models $p_U, p_B$. Based on the results we select one model combination to ablate for long-horizon performance and to study the impact of our model minimization approach in the face of distractors.

\begin{table}[t]
    \centering
    \caption{Simulation results with a comparison of class maps and different feature encodings in $0$-step inference. We bold-face the best performance in each row. We normalized the positional errors by the extent of the scenes before averaging them to account for different scene scales. The column $min.$ indicates the incurred error when a model is minimized. %
    }
    \vspace{0.2cm}
    \label{tab:map_performances_summarized}
    \small
    \begin{tabular}{ll|rrrr}
    \toprule
    Class & Pose &  \multicolumn{2}{c}{$\Delta \%$} & \multicolumn{2}{c}{$\Delta ^\circ$} \\
     map &  encoding & base & min. & base & min. \\
    \midrule
    \multirow[c]{5}{*}{$\M_I$} & $f_{quat}$ & 14.6 & 14.2 & 17.6 & \textbf{17.1} \\
     & $f_{mix}$ & 14.2 & 14.3 & 18.1 & 18.0 \\
     & $f_{euler}$ & 13.9 & 13.9 & 18.0 & 20.0 \\
     & $f_{AA}$ & 13.6 & 14.8 & 18.3 & 21.3 \\
     & $f_{\mathfrak{se3}}$ & 13.4 & 14.6 & 18.3 & 21.3 \\
     \midrule
    $\M_O$ & $f_{quat}$ & 5.8 & 5.9 & 17.6 & 18.2 \\
    \midrule
    \multirow[c]{2}{*}{$\M_U$} & $f_{quat}$ & 5.8 & 5.7 & \textbf{17.0} & 18.4 \\
                              & $f_{mix}$  & 5.6 & 5.6 & 20.4 & 20.7 \\
    \midrule
    \multirow[c]{2}{*}{$\M_O$} & $f_{\mathfrak{se3}}$ & 5.6 & 6.7 & 17.7 & 20.0 \\
          & $f_{mix}$ & 5.5 & 5.7 & 20.4 & 19.9 \\
    \midrule
    \multirow[c]{3}{*}{$\M_U$} & $f_{AA}$ & 5.5 & 5.7 & 18.8 & 21.1 \\
     & $f_{euler}$ & 5.5 & \textbf{5.5} & 18.0 & 19.7 \\
     & $f_{\mathfrak{se3}}$ & \textbf{5.4} & 5.8 & 17.1 & 17.8 \\
    \midrule
    \multirow[c]{2}{*}{$\M_O$} & $f_{euler}$ & \textbf{5.4} & \textbf{5.5} & 18.7 & 19.4 \\
     & $f_{AA}$ & \textbf{5.4} & \textbf{5.5} & 18.6 & 19.1 \\
    \bottomrule
    \end{tabular}
    \vspace{-0.5cm}
\end{table}

\begin{figure}
    \vspace{0.1cm}
    \centering
    \includegraphics[width=1.0\linewidth]{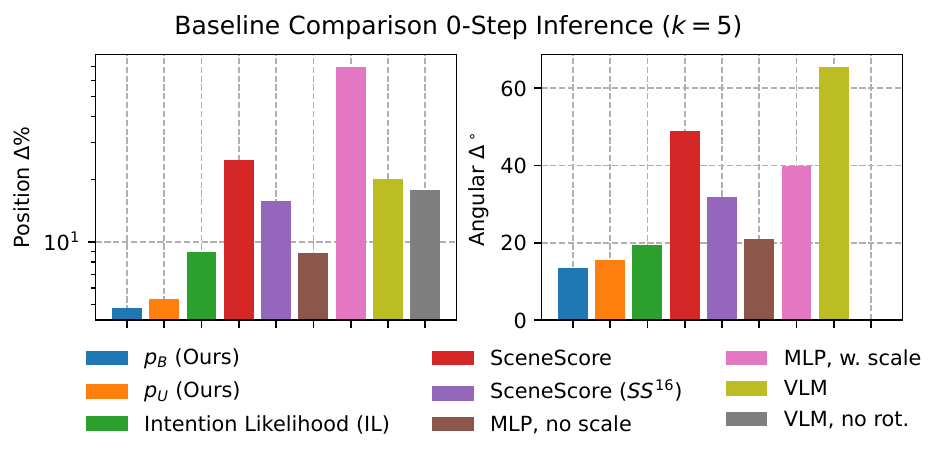}
    \caption{Results of our baseline comparison. For each method, we report the best-performing configuration. Our approach produces the lowest inference errors, followed by an MLP without object scaling information, Intention Likelihood, and Scene Score with 16 training samples \scenescorest. \change{The VLM performs slightly lower than \scenescorest and seems distracted by rotation inference.}}
    \label{fig:baseline_comparison}
    \vspace{-0.7cm}
\end{figure}

\begin{table*}[]
    \centering
    \caption{Comparison of bi-directional model $p_B$ using the \ifmix$f_{mix}$ and $f_{AA}$ pose representations\else $f_{AA}$ pose representation\fi{} against the three best baselines from \cref{fig:baseline_comparison} in the face of a growing number of distractors. 
    Columns $t < n$ show the mean error over a $n$-step inference. The lowest inference errors are highlighted per $d$-block and column. As the number of distractors increases, the minimization manages to eliminate distractor objects. The correlated rise of inference error and number of distractors indicates that neither method filters out all distractors. We note that Intention Likelihood (IL) is particularly sensitive to distractors, but also SceneScore's (\scenescorest) performance is impacted. In the right-most four columns we report the impact of the number of training samples. Given a larger number of samples, the benefit of our minimization scheme dissipates, while it demonstrates a particular effectiveness given fewer training samples.}
    \vspace{0.2cm}
    \label{tab:distractors}
\begin{tabular}{lcll|rrr|rrrcrrrr}
\toprule
Distractors & Minimized & Model & $f$ & \multicolumn{3}{c}{Position Error in $\Delta \%$} & \multicolumn{3}{c}{Angular Error in $\Delta ^\circ$} & & \multicolumn{4}{c}{$\Delta \%, t < 15$} \\
 &  &  &  & $t < 5$ & $t < 10$ & $t < 15$ & $t < 5$ & $t < 10$ & $t < 15$ & & $k = 3$ & $k = 4$ & $k = 5$ & $k=16$ \\
\midrule
\multirow[c]{7}{*}{$d = 0$} 
 &          & \scenescorest  &          & 18.2 & 18.5 & 18.5 & 38.0 & 39.5 & 40.1     &  &   -- &   -- &   -- & 18.5 \\
 &          & MLP & $f_{AA}$ & 10.0 & 10.2 & 10.2 & 27.7 & 30.6 & 31.4     &  & 13.5 & 11.3 & 10.2 & 7.3 \\
 &          & IL  &          & 9.6 & 9.6 & 9.6 & 23.5 & 25.1 & 25.6     &  &  9.1 &  9.4 &  9.6 & 8.7 \\
 & $\times$ & $p_B$ & $f_{mix}$ & 5.5 & 5.6 & 5.7 & 17.1 & 18.9 & 19.3     &  &  \textbf{6.6} &  5.9 &  5.7 & 4.7 \\
 & $\checkmark$ & $p_B$ & $f_{mix}$ & 5.4 & 5.5 & 5.6 & 16.4 & \textbf{18.4} & \textbf{19.0} &  &  \textbf{6.6} &  5.9 &  5.6 & 4.7 \\
 & $\times$ & $p_B$ & $f_{AA}$ & 5.3 & 5.5 & 5.5 & \textbf{16.7} & 18.6 & 19.1      &  &  \textbf{6.6} &  5.9 &  5.5 & 4.7 \\
 & $\checkmark$ & $p_B$ & $f_{AA}$ & \textbf{5.3} & \textbf{5.4} & \textbf{5.4} & \textbf{16.7} & 18.8 & 19.3  &  &  6.7 &  \textbf{5.8} &  \textbf{5.4} & \textbf{4.6} \\
 \cmidrule{2-10}\cmidrule{12-15}
\multirow[c]{7}{*}{$d = 1$} 
 &          & IL  &          & 35.1 & 34.9 & 34.9 & 57.4 & 58.5 & 58.9     &  & 30.1 & 36.6 & 34.9 & 56.1 \\
 &          & \scenescorest  &          & 27.9 & 28.6 & 28.7 & 48.4 & 50.3 & 50.6     &  &   -- &   -- &   -- & 28.7 \\
 &          & MLP & $f_{AA}$ & 13.7 & 13.8 & 13.8 & 40.1 & 44.2 & 45.2     &  & 16.7 & 14.5 & 13.8 & 9.5 \\
 & $\times$ & $p_B$ & $f_{AA}$ & 9.6 & 9.8 & 9.9 & 24.1 & 25.9 & 26.3      &  &  \textbf{6.5} &  8.5 &  9.9 & 4.8 \\
 & $\times$ & $p_B$ & $f_{mix}$ & 8.9 & 9.1 & 9.1 & 20.6 & 22.5 & 23.0     &  &  7.1 &  7.8 &  9.1 & \textbf{4.7} \\
 & $\checkmark$ & $p_B$ & $f_{AA}$ & 8.9 & 9.1 & 9.1 & 22.2 & 24.2 & 24.7  &  &  6.6 &  7.4 &  9.1 & \textbf{4.7} \\
 & $\checkmark$ & $p_B$ & $f_{mix}$ & \textbf{8.6} & \textbf{8.8} & \textbf{8.8} & \textbf{18.4} & \textbf{20.3} & \textbf{20.9} &  &  7.2 &  \textbf{6.8} & \textbf{8.8} & \textbf{4.7} \\
 \cmidrule{2-10}\cmidrule{12-15}
\multirow[c]{7}{*}{$d = 5$}
 &          & IL    &          & 60.2 & 59.5 & 59.3 & 99.4 & 99.6 & 99.7 &  & 56.1 & 59.8 & 59.3 & 80.8 \\
 &          & \scenescorest    &          & 36.1 & 36.2 & 36.1 & 92.4 & 94.0 & 94.0 &  &  -- &   -- &   -- & 36.1 \\
 & $\times$ & $p_B$ & $f_{AA}$ & 35.0 & 35.6 & 35.7 & 77.6 & 79.9 & 80.5 &  & 22.1 & 28.1 & 35.7 & 4.9 \\
 &          & MLP & $f_{AA}$ & 19.9 & 20.3 & 20.3 & 57.8 & 61.0 & 61.7                              &  & 30.9 & 24.9 & 20.3 & 12.5 \\
 & $\times$ & $p_B$ & $f_{mix}$ & 17.8 & 18.1 & 18.2 & 42.2 & 44.4 & 45.1                           &  & 13.8 & 19.9 & 18.2 & \textbf{4.7} \\
 & $\checkmark$ & $p_B$ & $f_{AA}$ & \textbf{12.8} & \textbf{13.0} & \textbf{13.0} & 26.8 & 28.9 & 29.3                        &  &  \textbf{9.9} &  8.9 & \textbf{13.0} & 4.8 \\
 & $\checkmark$ & $p_B$ & $f_{mix}$ & \textbf{12.8} & \textbf{13.0} & \textbf{13.0} & \textbf{25.3} & \textbf{27.3} & \textbf{27.9}                       &  & 11.4 &  \textbf{6.9} & \textbf{13.0} & \textbf{4.7} \\
\bottomrule
\end{tabular}
 \vspace{-0.3cm}
\end{table*}

\textbf{Results:} We present the results in Tab.~\ref{tab:map_performances_summarized} for 0-step inference. We find that the orthonormal class map $\M_O$ and $f_{AA}$ feature encoding achieve the most accurate result combining position and rotation. We note that the difference in positional error between the $\M_O$ and $\M_U$ is minor, while not using either scaling map incurs large errors. We note that the encoding $f_{quat}$ performs the worst out of all our pose encodings. Similarly, the automatically chosen pose representation $f_{mix}$ does not yield any benefit in $0$-step inference. We compare the performance of $p_U, p_B$ using the $\M_O$ category map and $f_{AA}$ pose representation and find $p_B$ to perform better on position ($5.3\% < 5.7\%$), and angle ($16.7^\circ < 21.0^\circ$). To evaluate our model's performance, we compare with SceneScore (SS)~\cite{kapelyukh2023scenescore} and the \textit{Intention Likelihood} (IL) model~\cite{welschehold2019combined}, \change{a VLM baseline~\cite{hurst2024gpt}}, as well as standard models. We obtain the implementation details of IL from \cite{abdo2017learning} and use the same hyperparameters as \cite{abdo2017learning}. For SS, we use the official code release and reduce our problem to 2D inferences on the XY plane. We train SS with the original training routine and hyperparameters on $k=16$ training samples, as this is the lowest sample count reported by the authors. \change{We present the VLM with overhead images of a current scene and a cropped image of the object to be placed. We prompt the model to state the pixel location where the new object should be placed. Please see the supplementary\footnote{Supplementary: \url{https://arxiv.org/abs/2411.03408}} for further details.}\ In addition to these works, we \change{also deploy an MLP on our scenes by stacking pose observations encoded as $f_{AA}$ and object scales and requiring it to infer the next object's pose. As MLPs require more training data}, we augment the original $k$ samples by applying random translations and rotations to the scene. We report metrics in~\cref{fig:baseline_comparison}. We include our Top-3 competitors (MLP, IL, \scenescorest) in our comparisons.

\begin{figure*}
    \vspace{4mm}
    \centering
    \footnotesize
    \includegraphics[width=\textwidth]{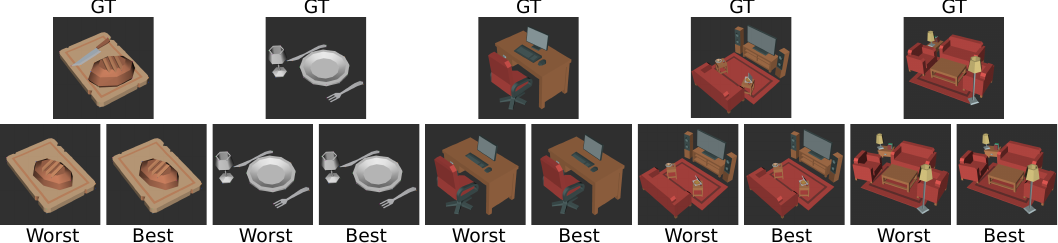}
    \caption{Renderings of scenes with highest $max$-step inference score, instead of lowest metric error. \textbf{GT}: Ground truth scene; \textbf{Best}: Best inference on this scene instance. \textbf{Worst}: Inference with the lowest joint score across all inference steps on the scene. The samples seem very close with no noticeable difference between them as a result of the refinement sampling. The scenes seem semantically plausible, indicating that model score is indicative of success. Yet, in the bread-cutting scene and the desk scene, thin objects such as a knife and keyboard get placed slightly below the supporting object's surface.
    }
    \label{fig:qualitative_inference}
    \vspace{-0.3cm}
\end{figure*}

We study both the long-horizon performance of our models and their sensitivity to distractor objects. In \cref{tab:distractors}, we show the average inference errors over longer inference sequences by computing the mean error incurred up to a step length of $t$. Additionally, we contrast the performances under a rising number of distractors. We do so by generating a set of up to $5$ distractor objects $d$ per scene instance and including them as observations in the model's fitting process. The distractors are distributed uniformly across the volume of the scene with an additional $50\%$ margin and possess uniformly sampled orientation and scale.
We note that inference errors remain steady over the length of an inference within one condition. As the number of distractors increases, our proposed minimization approach demonstrates its benefit. Yet, the rise in inference error overall does indicate that some distractor objects remain under consideration from the model. In \cref{tab:distractors}, we also show cumulative position error with respect to the number of training samples $k$. \change{While the most similar baseline IL, performs well in the noise-free condition with ample samples ($d=0,k=5$), its constant kernel density assumption becomes problematic in the noisier settings. Moreover, as the success of its inference is dependent on its training samples, it also suffers under scarcer training data. While the deep models (MLP, \scenescorest) are more robust to noise, the MLP's inference is impacted by decreasing training samples. Given the results in \cref{fig:baseline_comparison}, we do not ablate $k$ for \scenescorest. Interesting is that our model's performance improves with lower $k$ in $d>0$ setting}. This can be attributed to the regularization of the model size dependent on the number of training samples as described in \cref{sec:encoding_choice}. It seems the chosen hyperparameter of our minimization is more apt for less training data.
We present a qualitative impression of the best and worst inferences according to the model's own scores in \cref{fig:qualitative_inference}.
Despite these good results, we want to acknowledge that our model can only learn unimodal setups, while our baselines MLP, IL, and \scenescorest do not share this limitation.

\begin{figure*}
    \centering
    \includegraphics[width=\textwidth]{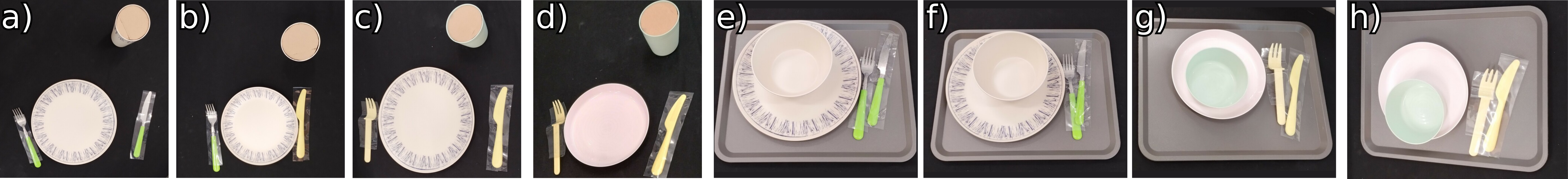}
    \caption{Eight examples of table settings by our robot. Our system is trained only with examples using real cutlery objects. Using our approach, the trained models transfer zero-shot to novel object instances, even though these vary significantly in their size. Using the training-free perception pipeline we describe, we are able to correctly associate object types despite stark visual differences. }
    \label{fig:real-results}
    \vspace{-0.3cm}
\end{figure*}

\subsection{Real Robotic Experiments: Table Setting}
\label{sec:robot-exp}
To evaluate our method on a real robot, we consider a table setting task. We implement a perception pipeline for detection and pose estimation of several intra-category instances of common tableware such as plates, bowls, cutlery, etc. Our pipeline infers object pose $\tf{W}{T}{o_t}$ and category map $\M_{o_t}$ from RGB-D data as shown in Fig~\ref{fig:pipeline}. We leverage several publicly available deep learning models for perception. We associate each class $c$ with a natural language label \ie \emph{fork} and use CLIP~\cite{radford2021learning} and the MaskCLIP technique~\cite{zhou2022extract} to extract the most relevant region in the RGB image. We feed the center of the highest scoring region to Segment Anything~\cite{kirillov2023segany}, which predicts a segmentation of the relevant object.
Similar to Goodwin~\textit{et~al.}~\cite{goodwin2023you}, we use DINOv2~\cite{oquab2023dinov2} to generate dense features for the pixels under the mask. Using back projection from the depth image, we obtain a 3D feature cloud $\F_{o}$. While~\cite{goodwin2023you} presents a method for category-level 6D pose estimation using such features, the stability of this approach is not sufficient for our task. Instead, we use the moments of our gathered points $\F_{o}$ to initialize a 6D pose. We apply a three-dimensional PCA transform from our categorical feature set $\F_{c_o}$ to $\F_{o}$ and compare the resulting features at the extremes of the longest moment of our point clouds. If the points match their opposite end better than the current one, we rotate the 6D pose around the shortest moment. The category map $\M_U$ is then identified from the ratios between the moments. In real robot experiments, due to noise in the depth image, we were unable to estimate an $\M_O$ mapping reliably and used $\M_U$.
We used a Kawada Robotics Nextage dual-armed robot to perform our table setting experiments. Our perception pipeline requires an average of \SI{1050}{\milli\second} to extract an object and match its pose and up to 7.5 GB of VRAM on a NVIDIA RTX 4060 Ti with 16 GB of VRAM. As we only invoke the pipeline once per placement, we consider this latency acceptable.

\textbf{Experimental Setup:} For this experiment, we set up two tables as shown in Fig~\ref{fig:pipeline}. On one table, the objects are arranged randomly. On the other table is a single static object to initiate the relative pose prediction. The robot is given a list of objects and the order they must be placed in. However, object detection, grasping, predicting placement pose, and placement are all performed autonomously. First, the robot examines the table of placement objects, detects the query object, and grasps it. The robot then examines the table of static objects, locating all the current static objects, and predicts where to place the new object, and then places it. The process continues until there are no more placement objects. We train models for these tasks using four demonstrations, each with only the real tableware. As some objects, such as plates and cups, are rotationally invariant, we minimize the variance in rotations for these objects among the demonstrations. 

\textbf{Results:} In Fig.~\ref{fig:real-results}, we show eight examples of tables set by the robot. This includes novel object instances and combinations of different cutlery sets. The robot adapted object placement for different-sized objects. For example, in Fig.~\ref{fig:real-results}~g) in spite of the smaller children's plate, the robot still places the fork and knife close to the plate. We also randomized the initial placement of the tray, and the robot inferred that the whole scene should be rotated. We found the pose prediction to be very robust. Most failures in placement were due to errors in the perception system, such as sensor noise in the depth measurement. %

In an effort to better understand our method's performance, we surveyed 111 people, asking them to rate thirty images of table settings done by our system on a 10-point scale. We included 10 images of table settings done by humans to establish a baseline. We compared two scenes, one without a tray (similar to Fig.~\ref{fig:real-results} a-d) and one with a tray (Fig.~\ref{fig:real-results} e-h). Scaling the average robot scores by average human scores, the robot was rated 73.3\% as good as humans for setting tables without the tray. With the tray, this reduced to 62.2\%. This can be attributed due to the respondents rating the human tray setting higher than the table setting. We suspect the presence of the tray led respondents to rate the human settings more highly as the cutlery was better aligned with the tray. 

\section{Conclusion}

In this work, we presented an approach for efficient learning of relative object placement poses with intra-categorical transfer. We achieved this by introducing a mapping from observed objects to canonical class features, enabling transfer to unseen instances of different poses and scales. In simulated evaluations, we identify that a bi-directional model with an encoding of poses as positions and rotation vectors performs the best quantitatively, and we find that our model minimization approach is successful at removing distractors. We demonstrated that our approach can be deployed on real robotic systems to set tables with different, unseen object instances of varied scales. In human evaluations, our method was rated as good as 73.3\% compared to a human table setting baseline. We view this approach as a successful step towards efficient learning of object placements from demonstrations.

Going forward, we would like to consider including multi-modal models and exploring further pose encoding options, such as spherical coordinates, in our approach. The formulation of our approach lends itself to the inclusion of multi-modal models, but determining the number of modes can be challenging~\cite{figueroa2018physically}. Further, we would like to consider a different type of class mappings. Our chosen linear projections work but do not consider the affordances of objects, implying that these are always located similarly. This would make it difficult to reliably place a key in a keyhole on a door. Learning relative poses of feature points might be more useful, but may also require more data, \ie as done in~\cite{gao2023kvil}.

\section*{Acknowledgment}
We would like to thank our colleague Imen Mahdi for offering her expertise and help in creating the VLM-baseline.
In addition, we thank \href{https://kenney.nl}{kenney.nl} for their free 3D assets.

\bibliographystyle{IEEEtran}
\bibliography{sources}  %

\begin{center}

\Large{\bf- Supplementary Material -}\\
\end{center}

In this Supplementary, we provide additional details on our LLM-baseline, and implementation details of the implementation of our real system and the conducted Human survey.

\subsection{LLM-Baseline}
\label{apx:llm_baseline}

In our experiments, we use GPT-4o~\cite{hurst2024gpt} through the API provided by OpenAI. We present the model with an overhead image of the scene it is meant to complete and a cropped view of the object it is asked to place, as well as examples of completed scenes from the training dataset. Our prompt includes the names of the other objects in the scene and their pixel locations. The prompt then queries the model for a location to place the next object at in pixel coordinates. 
To compare these inferences with those of our other baselines, we project the given coordinates back to the ground truth depth at which our placement object is located and measure the Cartesian error. In a second condition, we change the prompt to display the placement objects at a standardized rotation and ask the model to also infer a rotation in degrees. We run each inference 5 times and average the results which yields the ones we report in~\cref{fig:baseline_comparison}. We share the prompts we used in \cref{apx:tab:vlm_prompts}. Notably, our prompt only includes one image of a whole scene as \textit{training} image. We also attempted to present the model with the actual training examples, but it would always repeat one of the previous answers. 

\begin{table*}[t]
    \centering
        \caption{Prompts used for the VLM baseline. We use a constant system prompt, followed by an introduction of the scenario, accompanied by an image of the scene. This is then followed by a description of object locations and an image of the current scene. Finally, we ask the model to predict the location of the object to be placed and show it an image of this object. The prompt differs slightly depending on whether the model is also asked to infer rotation.}
    \label{apx:tab:vlm_prompts}
    \begin{tabular}{p{0.07\linewidth} p{0.1\linewidth} | p{0.5\linewidth} | p{0.2\linewidth}}
    \toprule
    \multicolumn{2}{l|}{Prompt Type}  & Prompt & Example Image \\
    \midrule
     \multicolumn{2}{l|}{System Prompt}  & You are a helpful AI assistant. You will help place items in a scene. RESPECT THE USER SETUP PREFERENCE WHEN PLACING OBJECTS. \\
     \midrule
     \multicolumn{2}{l|}{Task Introduction} & Here is an example of a final setup of the scene. In the following messages, you should respect the setup preference of the user. & \raisebox{-\totalheight}{\includegraphics[width=0.9\linewidth]{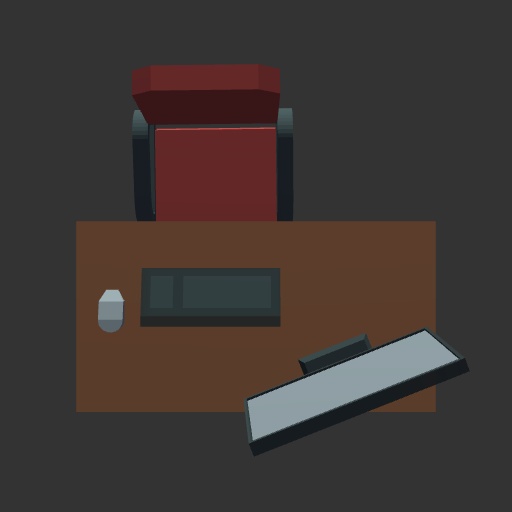}} \\
     \midrule
     \multirow[c]{24}{*}{\parbox{0.9\linewidth}{Task Prompt}} & \multirow[c]{16}{*}{\parbox{\linewidth}{Only positional inference}} & Here is a top view of the scene. Current Objects placed are: [OBJECT\_1:x=PX\_X,y=PX\_Y, ..., OBJECT\_N:x=PX\_X,y=PX\_Y].
      & \raisebox{-\totalheight}{\includegraphics[width=0.9\linewidth]{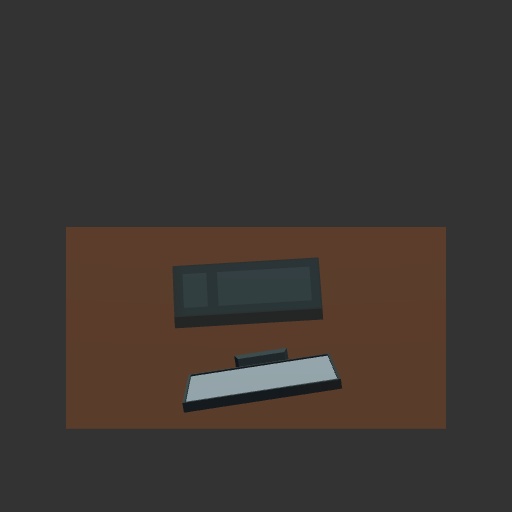}} \\
     \cmidrule{3-4}
     & & Here is a top view of the object to place at the correct scale and the correct rotation. At which pixel location in the previous scene should the center of the object: OBJECT (center represented by the pink pixel) be placed. Predict the x\_pixel,y\_pixel positions of the frame origin. Return only the format "x=,y=" and nothing else. Always reply with this format even when not certain. & \raisebox{-\totalheight}{\includegraphics[width=0.9\linewidth]{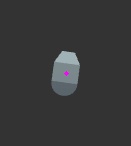}} \\
    \cmidrule{2-4}
     & \multirow[c]{16}{*}{\parbox{\linewidth}{Positional and rotational inference}} &  Here is a top view of the scene. Current Objects placed are: [OBJECT\_1:x=PX\_X,y=PX\_Y,r=YAW, ..., OBJECT\_N:x=PX\_X,y=PX\_Y,r=YAW].
      & \raisebox{-\totalheight}{\includegraphics[width=0.9\linewidth]{figures/vlm/desk_inf_scene.jpg}} \\
     \cmidrule{3-4}
     & & Here is a top view of the object to place at the correct scale. Where and at which angle in the previous scene should the frame (marked by red x axis and green y axis) of the object: OBJECT be placed. Predict the x\_pixel,y\_pixel positions of the center of the object and the yaw rotation angle r (between -180 and 180) with rotation being positive for counter clockwise. Return only the format 'x=,y=,r=' and nothing else. Always reply with this format even when not certain. & \raisebox{-\totalheight}{\includegraphics[width=0.9\linewidth]{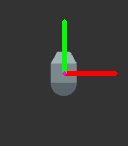}} \\
    \bottomrule
    \end{tabular}
\end{table*}

\subsection{Invariance Preprocessing}
\label{apx:invariance_processing}

In \cref{sec:robot-exp}, we indicate a preprocessing step that we use to minimize the effect of rotational invariance in our real training observations. This preprocessing step is as follows: for the categories of \textit{plate}, \textit{bowl}, \textit{cup}, and \textit{tray}, we identify an axis of rotational invariance and for the tray, an additional \textit{step increment} of invariance, \ie the tray is only invariant in $180^\circ$ increments.

Given a set of $K$ training observations, we use the location of the final object $o_n$ as a reference point. We use the training example $k=1$ to establish a canonical observation. We do so by calculating the bearing angle of $o_{1,n}$ in the plane orthogonal to the axis of invariance of the invariant object $\hat{o}_{1,i}$ as $\alpha_{1,i}$. For all $k>1$ we then rotate $\hat{o}_{k,i}$ such that $\alpha_{k,i} = \alpha_{1,i}$ for $o_{k,n}$. In the case of the tray, the rotation is rounded to $180^\circ$ increments.
As the final object $o_n$ can also be invariant, we apply the same procedure to it by using $o_1$ as a point of reference.
During inference, no such procedure is needed, as the invariant objects are aligned to the inferred poses.

This technique for minimizing the effect of invariance is simple and efficient. It would be interesting to use detected annotations of invariance instead of handcrafted ones, but this is outside of the scope of this work. In addition, this process only works in the case of unimodal observations. For multimodal observations, the invariance minimization and model fitting would have to happen in tandem.

\subsection{Motion Generation}
\label{apx:motion_generation}

To manipulate our objects, we generate trajectories using a very simple IK-scheme. We record the relative object-endeffector transformation at the time of object pick-up and use this transform in combination with the inferred pose to define a pose goal for the end-effector. This goal is set so that the object is a few centimeters above its goal position. From this pose, we generate a pre-pick pose and a post-pick pose for our end-effector. Starting from the initial robot pose, we solve the IK for this series of poses, initializing the subsequent IK-problem with the preceding one’s solution to ensure minimal joint-space deltas. We then execute this series in joint space, generating a smooth trajectory between two points using cubic splines.

\subsection{Human Evaluation Survey}

\begin{figure*}
    \centering
    \begin{subfigure}[t]{\linewidth}
        \centering
        \includegraphics[height=5cm]{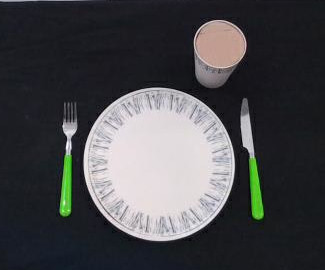}
        \includegraphics[height=5cm]{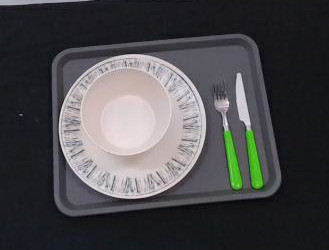}
        \caption{Examples shown do participants and demonstrators as desirable.}
        \label{subfig:scene_demos}
    \end{subfigure}
    \vspace{1mm}

    \begin{subfigure}[t]{\linewidth}
    \includegraphics[width=\linewidth]{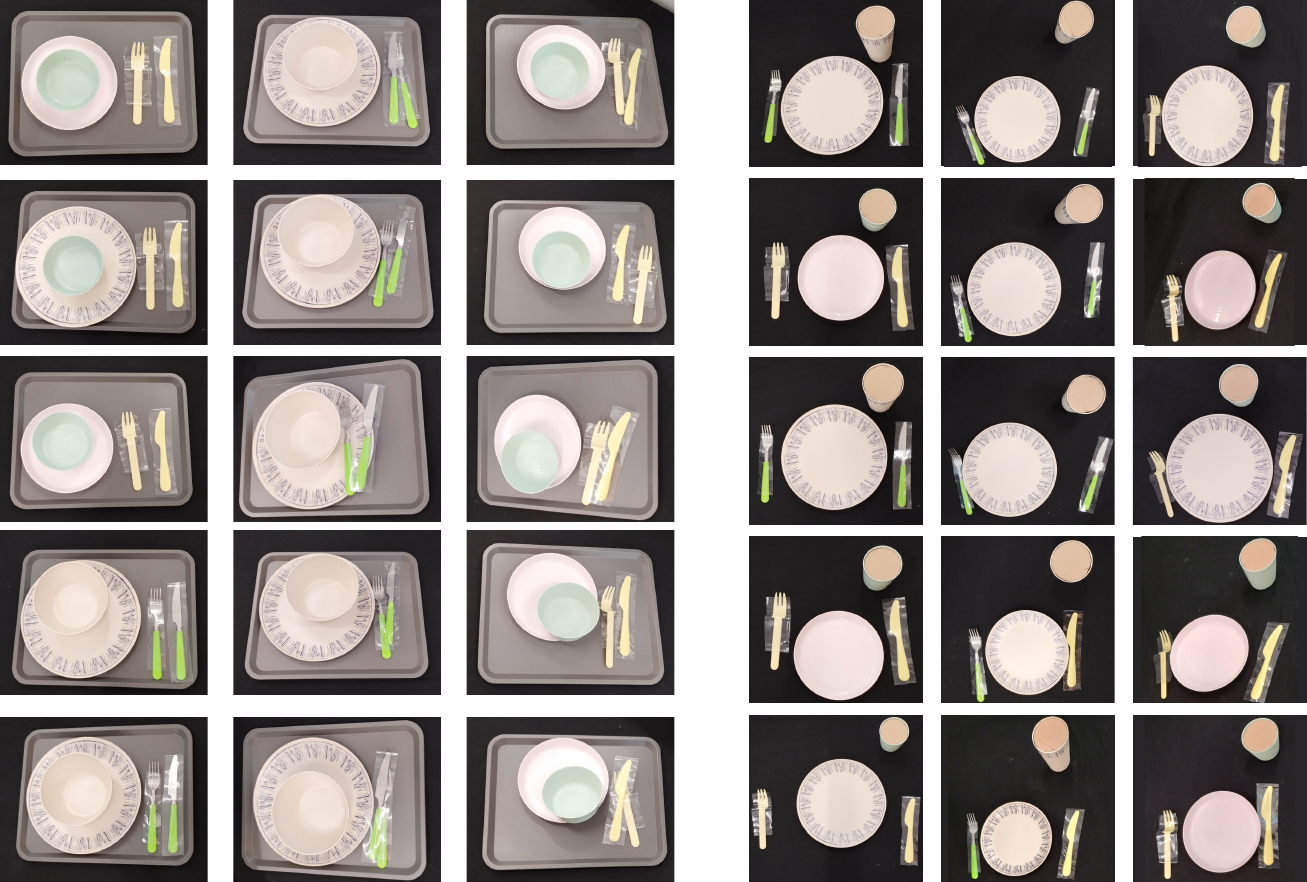}
    \caption{Scenes presented to participants. The first column of each scenario was set by humans, the latter columns were set by our the robot.}
    \label{subfig:all_scenes}
    \end{subfigure}
    \caption{The images in \cref{subfig:all_scenes} were presented to participants to be rated on a scale of $0-10$ from \textit{unacceptable} to \textit{perfect}, given the images \cref{subfig:scene_demos} as prior.}
    \label{fig:study_images}
\end{figure*}

In our human evaluation survey, as described in \cref{sec:robot-exp}, we present participants with 30 images of set tables and trays, 10 of which are set by humans. For each scenario, we show one example image captured from the robot's perspective, which matches the robot's demonstration data. These examples were also shown to our colleagues as prior for setting the tray and the table. All images are shown in \cref{fig:study_images}.

\end{document}